\documentclass[runningheads]{llncs}

\usepackage[T1]{fontenc}

\usepackage{graphicx}
\graphicspath{{./pics/}{./}}

\usepackage{amsmath}
\usepackage{algorithm}
\usepackage{algorithmic}
\usepackage{subcaption}
\usepackage{booktabs}

\usepackage{url}
\usepackage[allcolors=black,hidelinks]{hyperref}

\usepackage[absolute]{textpos}
\usepackage{orcidlink}
\hypersetup{hidelinks}

\begin{document}

\begin{textblock}{122}(47.5,240)
\noindent \scriptsize 
This preprint has not undergone peer review (when applicable) or any post-submisslon improvements or corrections. The Version of Record of this contribution is published in Networked Systems, and is available online at \url{https://doi.org/10.1007/978-3-031-17436-0_20}\\

\noindent The original article appeared as: Janne Alatalo, Joni Korpihalkola, Tuomo Sipola and Tero Kokkonen. “Chromatic and spatial analysis of one-pixel attacks against an image classifier.” In: Networked Systems. NETYS 2022. Ed. by Mohammed-Amine Koulali and Mira Mezini. Vol. 13464. Lecture Notes in Computer Science. Cham, Switzerland: Springer, 2022, pp. 303--316. DOI: \href{https://doi.org/10.1007/978-3-031-17436-0_20}{10.1007/978-3-031-17436-0\_20}
\end{textblock}

\title{Chromatic and spatial analysis of one-pixel attacks against an image classifier}

\titlerunning{Chromatic and spatial analysis of one-pixel attacks}

\author{
Janne Alatalo\orcidlink{0000-0001-5515-4419} \and 
Joni Korpihalkola\orcidlink{0000-0001-6434-1240} \and 
Tuomo Sipola\orcidlink{0000-0002-2354-0400} \and 
Tero Kokkonen\orcidlink{0000-0001-9988-6259}
}

\authorrunning{J.\ Alatalo et al.}

\institute{
Institute of Information Technology, JAMK University of Applied Sciences,\\ Jyv{\"a}skyl{\"a}, Finland
\email{\{janne.alatalo,joni.korpihalkola,tuomo.sipola,tero.kokkonen\}@jamk.fi}
}

\maketitle

\begin{abstract}

\noindent
One-pixel attack is a curious way of deceiving neural network classifier by changing only one pixel in the input image. 
The full potential and boundaries of this attack method are not yet fully understood. 
In this research, the successful and unsuccessful attacks are studied in more detail to illustrate the working mechanisms of a one-pixel attack created using differential evolution. 
The data comes from our earlier studies where we applied the attack against medical imaging. We used a real breast cancer tissue dataset and a real classifier as the attack target. 
This research presents ways to analyze chromatic and spatial distributions of one-pixel attacks. In addition, we present one-pixel attack confidence maps to illustrate the behavior of the target classifier. 
We show that the more effective attacks change the color of the pixel more, and that the successful attacks are situated at the center of the images. 
This kind of analysis is not only useful for understanding the behavior of the attack but also the qualities of the classifying neural network. 

\end{abstract}

\section{Introduction}
\label{sec:intro}

The use of Artificial Intelligence (AI), including sub-branches Machine learning (ML) and Deep Learning (DL), is continuously increasing as support for decision making in automated image analysis of medical imaging~\cite{Zhou_2021,Latif-2019}. One enabler for such evolution is that there is the abundance of available data for research and development activities in the medical domain~\cite{Sasubilli-2020}. However, from the cyber security standpoint, this evolution fosters attack surface, and it should be realized that new technologies attract malicious actors and especially medical domain can be seen as a valuable target to gain profit by causing disruptions. It is noticeable that most of the medical data has sensitive nature. For example, Europol has announced that during the ongoing COVID-19 pandemic, the pandemic-themed cybercrime activities and campaigns are also targeted to healthcare organizations.~\cite{Europol_2020}. Newaz et al.\ propose an adversarial attack against ML enabled smart healthcare system~\cite{Newaz_2020}. Attacks against new technologies might induce harmful effects: considerable time to recover, mistrust against AI-based models and even fear of misdiagnosis. It is noticeable that Internet of Things (IoT) devices have a remarkable role in the healthcare~\cite{Bharadwaj_2021} and there are known security issues with IoT. Several AI models are in risk for adversarial attacks~\cite{Watson_2021} Liu et al.\ introduce and summarize the DL associated attack and defense methods~\cite{Liu_2021}, while Qayyum et al.~\cite{Qayyum_2021} introduce methods to warrant secure ML for healthcare. Integrity and unauthorized usage of medical image data is important when considering attacks against AI based medical imaging. In that sense, Kamal et al.\ proposed image encryption algorithm for securing medical image data~\cite{Kamal_2021}.

One-pixel attack is an adversarial method that changes just one pixel in an image to cause misclassification~\cite{su2019one}. 
However, its sensitivity to change and effectiveness are not fully understood. 
A few methods have been proposed to visualize the effect of one-pixel attacks. Wang et al.\ propose pixel adversarial maps and probability adversarial maps~\cite{wang2020visualizing}. Vargas et al.\ go further, and use internal information from the neural network model to create propagation maps to show the influence of one-pixel attacks through convolution layers.~\cite{vargas2019understanding}


In this study, we provide tools to understand the behavior of a neural network classifier targeted by the one-pixel attack. 
Our present analysis is a natural extension to our prior studies related to the attack method. Earlier, we have introduced a list of methods to fool artificial neural networks used in medical imaging~\cite{Sipola_2020}. One-pixel attack appeared to be a comprehensive and realistic attack vector, so we decided to further investigate it as a conceptual framework in the medical imaging domain~\cite{Sipola_2021}. When the concept and usability of the attack were understood, we succeeded to implement the technical one-pixel attack against real neural network models used in medical imaging~\cite{korpihalkola2020onepixel}. That first technical attack was a success, but the pixel changes in the images were quite easily observable by a human. It seemed that the attack was not realistic or comprehensive for real-world attackers, so we decided to further develop the attack methodology~\cite{Korpihalkola_2021}. 

The rest of the paper is organized as follows. First, data source and analysis methods are introduced in section~\ref{sec:methods}. Results of chromatic, spatial and periodicity analysis are presented in section~\ref{sec:results}. Finally, the study is concluded with final discussion and future research topics in section~\ref{sec:conclusion}.

\section{Methods}
\label{sec:methods}

\subsection{Data source}
\label{sec:data}

In our previous publications we introduced how an artificial neural network image classifier model could be fooled by changing only one pixel in the input image~\cite{korpihalkola2020onepixel,Korpihalkola_2021}. Those studies targeted IBM CODAIT MAX breast cancer detector which uses a modified ResNet-50 model~\cite{ibmmax}. The model is an open-source convolutional neural network classifier predicting the probability that the input image contains mitosis. The previous studies used a pretrained version of the model that was trained using the TUPAC16 breast cancer dataset~\cite{tupac16,tupacpaper}. We use the same model in this research. 

The study used the one-pixel attack to find adversarial images that would make the model predict wrong results for the input images~\cite{su2019one}. This method uses differential evolution optimization, where a population of breast cancer images is attacked by randomly choosing one pixel and randomly changing the pixel's colors to new values. The color values are mutated until the lowest confidence score is achieved for the breast cancer image. The method efficiently finds possible one-pixel changes to the image that changes the prediction outcome.

The targeted model can be fooled in two ways. If the model predicts strong probability of mitosis for the input image, then the one-pixel attack is used to find the pixel that lowers the predicted mitosis probability when the pixel color is changed (\textit{mitosis-to-normal}). The other way to fool the model is to try to increase the predicted mitosis probability when the model predicts low mitosis confidence score for some input image (\textit{normal-to-mitosis}). The study explored both possible cases of fooling the model.
The study concluded that both \textit{mitosis-to-normal} and \textit{normal-to-mitosis} attacks are possible, but of those two, \textit{mitosis-to-normal} attacks are considerably easier to carry out.

The dataset used in this study contains the one-pixel attack results from the previous study~\cite{korpihalkola2020onepixel}, and information of the attacked image, such as the attack pixel's location in the image and the nearby neighboring pixels' color values of the attacked pixel. Attacks that were considered successful were selected from the dataset, \textit{mitosis-to-normal} attacks, where the confidence score of the neural network was decreased to less than $0.9$, was included. In \textit{normal-to-mitosis} attacks, if the attack in the experiment caused to the confidence score to rise above 0.1, the attack was included for visualizations. Using these filters, 3,871 \textit{mitosis-to-normal} attacks and 319 \textit{normal-to-mitosis} attacks were used as a visualization dataset.

\subsection{One-pixel attack confidence map computation}
\label{sec:bruteforce}

In addition to analyzing the results from our previous paper, we also carried out additional tests for some of the dataset images by brute forcing a subset of all possible attack vectors for the images, producing a one-pixel attack confidence map. This gave us a clearer view how the successful attack vectors were positioned in individual images. The brute force computation was conducted on a few handpicked images that were chosen based on our previous paper results in a way that we had successful and failed examples of both \textit{mitosis-to-normal} and \textit{normal-to-mitosis} attack types.

This research used color images, hence each pixel has three color channels and the color value for each channel has a value between \(0 - 255\). This means that the total number of possible colors for a single pixel is \(16,777,216\). The images were \(64 \times 64\) pixels in size, so the total number of all possible attack vectors is \(68,719,476,736\) for a single image. We concluded that computing all possible vectors for the images is not worthwhile; therefore, we settled on a subset of all possible attack colors. The selected set of colors \(C\) (\ref{formula:bruteforced-colors}) was generated by taking every fifth color value for each channel and taking all their color combinations. In the equation, \(r\), \(g\) and \(b\) are the red, green, and blue color channels:

\begin{equation}
  C = \{ (r, g, b) \mid r, g, b \in \{ 0, 5, 15 \ldots 255 \} \}.
  \label{formula:bruteforced-colors}
\end{equation}

Even when the brute forced colors were reduced to the set \(C\), there was still \(140,608\) different colors for a single pixel, meaning that the total number of attack vectors for a single image was still \(575,930,368\). With that many images we could not use the Docker containerized version of the model that was used in our previous study over the HTTP API, because the containerized version of the model does not support GPU computation or image batching. We overcame this problem by deploying the model to our computation server without the containerization layer and implementing a highly efficient GPU accelerated data-pipeline that implemented the one-pixel modifications on GPU without needing to continuously copy the images between CPU and GPU memory. With this setup computing the \(575,930,368\) attack vectors for one image took about 5 hours on our computation server using one Nvidia Tesla V100 GPU.

The results of the brute force attack vector analysis were reduced to minimum, maximum and average score values for each pixel coordinate.

Let \(I_{x,y}\) be the set of all modified images where pixel coordinate \((x, y)\) value is replaced with color value \(c \in C\) in the image under brute force computation. Let \(f\) be the model that predicts the score for the images. The results of the brute force attacks were processed with method described in Equation~\ref{formula:brute-force-processing} and Algorithm \ref{alg:brute-force-processing}.

\begin{equation}
  \begin{split}
    s_{max}(x, y) = \textrm{max}(\{ f(i) \mid i \in I_{x,y} \}) \\
    s_{min}(x, y) = \textrm{min}(\{ f(i) \mid i \in I_{x,y} \}) \\
    s_{avg}(x, y) = \textrm{avg}(\{ f(i) \mid i \in I_{x,y} \}) \\
  \end{split}
  \label{formula:brute-force-processing}
\end{equation}

\begin{algorithm}
  \caption{Brute force results processing algorithm}
  \label{alg:brute-force-processing}
  \begin{algorithmic}
    \STATE{$maxscores \leftarrow \textnormal{ARRAY}[64][64]$}
    \STATE{$minscores \leftarrow \textnormal{ARRAY}[64][64]$}
    \STATE{$avgscores \leftarrow \textnormal{ARRAY}[64][64]$}
    \FOR{$x \leftarrow 0$ to $63$}
      \FOR{$y \leftarrow 0$ to $63$}
        \STATE{$maxscores[x][y] \leftarrow s_{max}(x, y)$}
        \STATE{$minscores[x][y] \leftarrow s_{min}(x, y)$}
        \STATE{$avgscores[x][y] \leftarrow s_{avg}(x, y)$}
      \ENDFOR
    \ENDFOR
  \end{algorithmic}
\end{algorithm}

\section{Results}
\label{sec:results}

\subsection{Chromatic analysis}
\label{subsec:chromatic}

The difference between color values of two different pixels was measured by root mean square error (RMSE).

\[ h(\mathbf{x}) = \sqrt{\frac{(c_r - c_{r\mu})^{2} + (c_g - c_{g\mu})^{2} + (c_b - c_{b\mu})^{2}}{3}}, \]

\noindent
where $c_r$, $c_b$, $c_g$ are the color values of the attack vector and $c_{r\mu}$, $c_{g\mu}$, $c_{b\mu}$ are the means of the attack vector's surrounding pixels' color values. All values were scaled within the range $[0,1]$.

\begin{figure}[htp]
    \centering
    \includegraphics[width=0.9\textwidth]{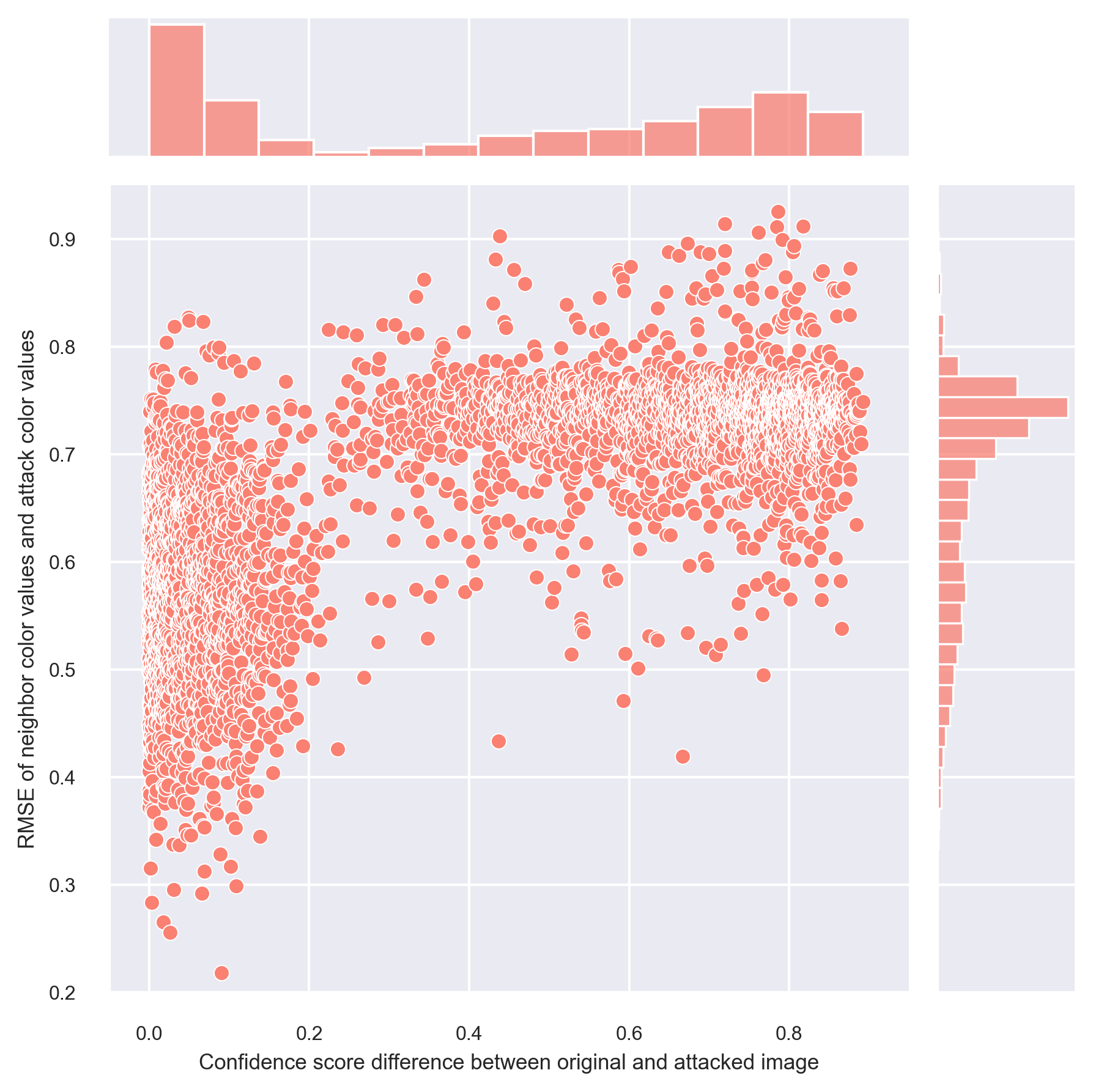}
    \caption{Scatter plot of error function values between the attack pixel color values and neighboring pixel color values. Notice the vertical cluster at low error values and horizontal cluster at higher error values.}
    \label{fig:mitosis-to-normal-scatterplot}
\end{figure}

When the attacks managed to fool the neural network, the error function values were high in \textit{mitosis-to-normal} attacks, as can be observed from Figure~\ref{fig:mitosis-to-normal-scatterplot}, which shows one vertical and one horizontal cluster. This indicates that the attacks which managed to lower confidence score the most had pixel color values noticeably different from the surrounding colors. The positioning of the attack pixel also matters, since some attacks had a higher color difference between neighboring pixels and still did not manage to lower the confidence score by more than $0.2$. 

\begin{figure}[htp]
    \centering
    \includegraphics[width=0.9\textwidth]{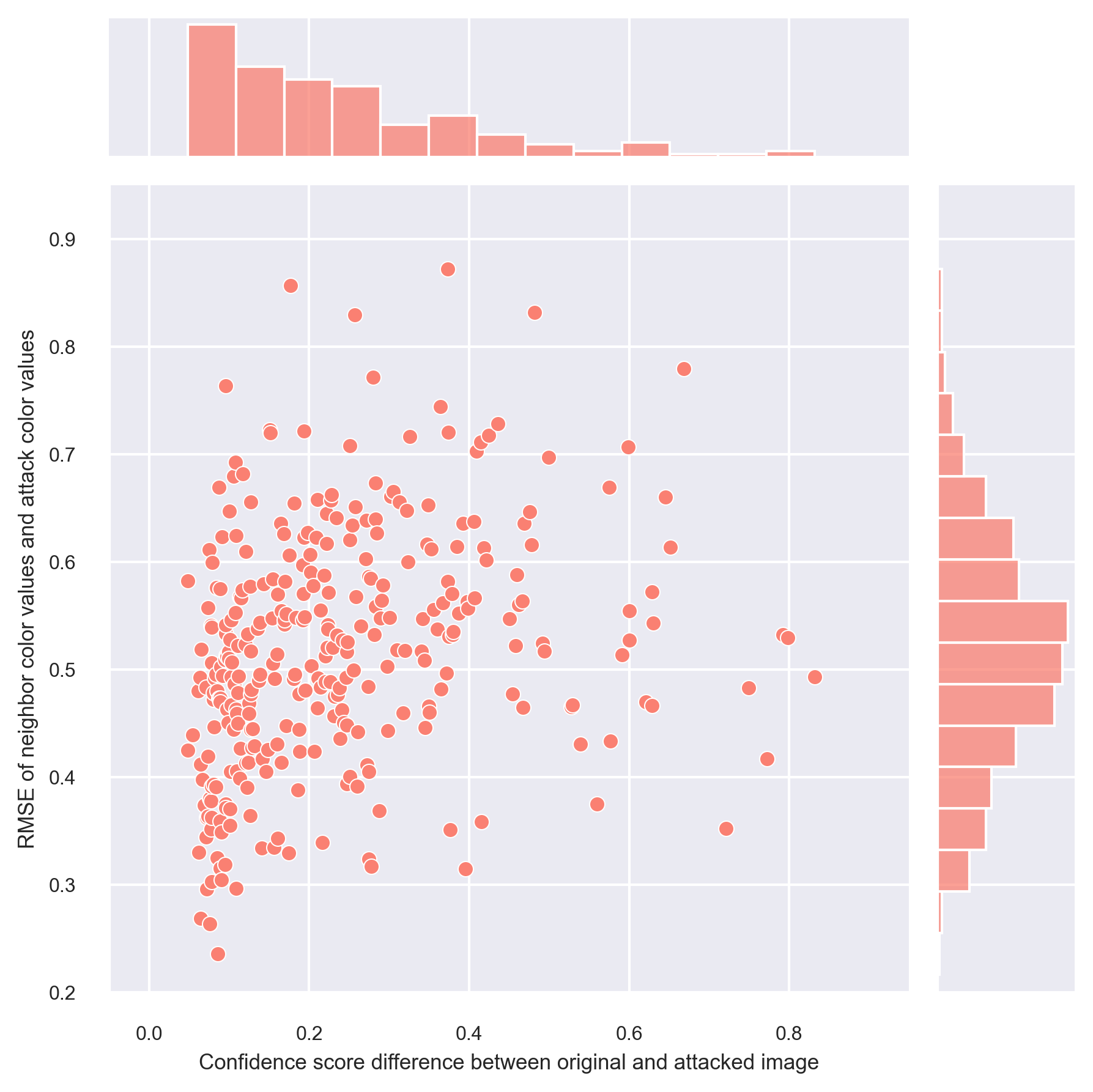}
    \caption{Scatter plot of error function values between the attack pixel color values and neighboring pixel color values.}
    \label{fig:normal-to-mitosis-scatterplot}
\end{figure}

In \textit{normal-to-mitosis} attacks the error values were lower than in \textit{mitosis-to-normal} attack, as can be seen in Figure~\ref{fig:normal-to-mitosis-scatterplot}, which shows no clusters; instead, the dots are more evenly distributed between the lower X axis values. 

\subsection{Spatial analysis}
\label{subsec:spatial}

Mean, median and standard deviation numerical measures were calculated for the attacks. In the Table~\ref{tab:mitosis-statistical-values}, the X and Y mean and median indicate that the attacks were mostly located at the center of the 64 by 64 pixels images. Meanwhile, the color values of red and green were near the maximum value of 255, while blue values were lower with higher standard deviation compared to red and green.

\begin{table}[htp]
    \centering
    \caption{Statistical measures for \textit{mitosis-to-normal} attacks ($N=3871$)}
    \label{tab:mitosis-statistical-values}
    \vspace{4pt}
    \begin{tabular}{l c c c c c}
    \toprule
                       & X     & Y     & Red    & Green  & Blue \\
    \midrule
    Mean               & 32.40 & 29.30 & 231.14 & 227.24 & 67.07 \\
    Median             & 32    & 30    & 255    & 255    & 37    \\
    SD                 & 8.2   & 8.59  & 41.62  & 45.99  & 77.85 \\
    \bottomrule
    \end{tabular}
\end{table}

In normal images, the statistical measures listed in Table~\ref{tab:normal-statistical-values} show that the attack vector is mostly again located at the center of the image, while there is much greater variation in red, green and blue color values, with a standard deviation between 90 and 100 in all of them.

\begin{table}[htp]
    \centering
    \caption{Statistical measures for \textit{normal-to-mitosis} attacks ($N=319$)}
    \label{tab:normal-statistical-values}
    \vspace{4pt}
    \begin{tabular}{l c c c c c}
    \toprule
                       & X     & Y     & Red    & Green  & Blue \\
    \midrule
    Mean               & 31.55 & 31.15 & 145.28 & 153.31 & 124.29 \\
    Median             & 32    & 32    & 154    & 168    & 129    \\
    SD                 & 10.54 & 10.77 & 92.62  & 93.04  & 99.53  \\
    \bottomrule
    \end{tabular}
\end{table}

The statistical measures show that the dataset is most likely preprocessed in such a manner that the features used by the neural network to classify an image to either mitosis or normal class are located in the center of the image. Higher red and green color values were the key in fooling the neural network in both attacks, while blue color values were closer to zero or in the middle of the color range. In the TUPAC16 dataset, the mitosis activity was low in color range, so the neural network might be fooled by values in the higher color range.

\subsection{Periodicity analysis}

\begin{figure}[htp]
    \centering
    \includegraphics[width=0.9\textwidth]{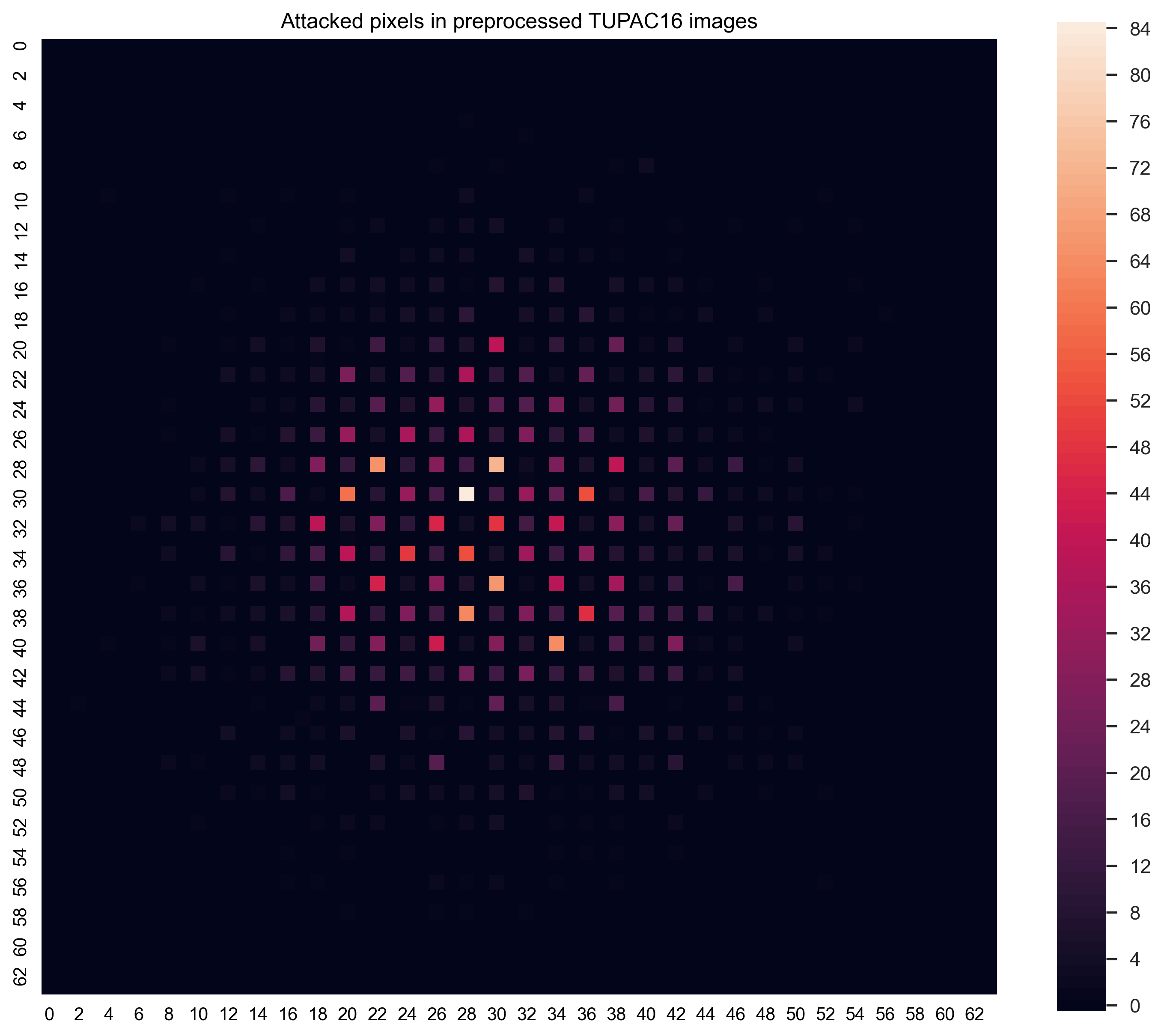}
    \caption{A heatmap of attack placements in images. Notice the checkerboard pattern at the center.}
    \label{fig:mitosis-to-normal-heatmap}
\end{figure}

The targeted model is a neural network with convolution layers, which shift through the input image in smaller windows and step to the right in steps.
To check for biases in the convolutional model, the best attack locations for all the target images is visualized in a heatmap in Figure~\ref{fig:mitosis-to-normal-heatmap}. 

\begin{figure}[htp]
    \centering
    \includegraphics[width=0.9\textwidth]{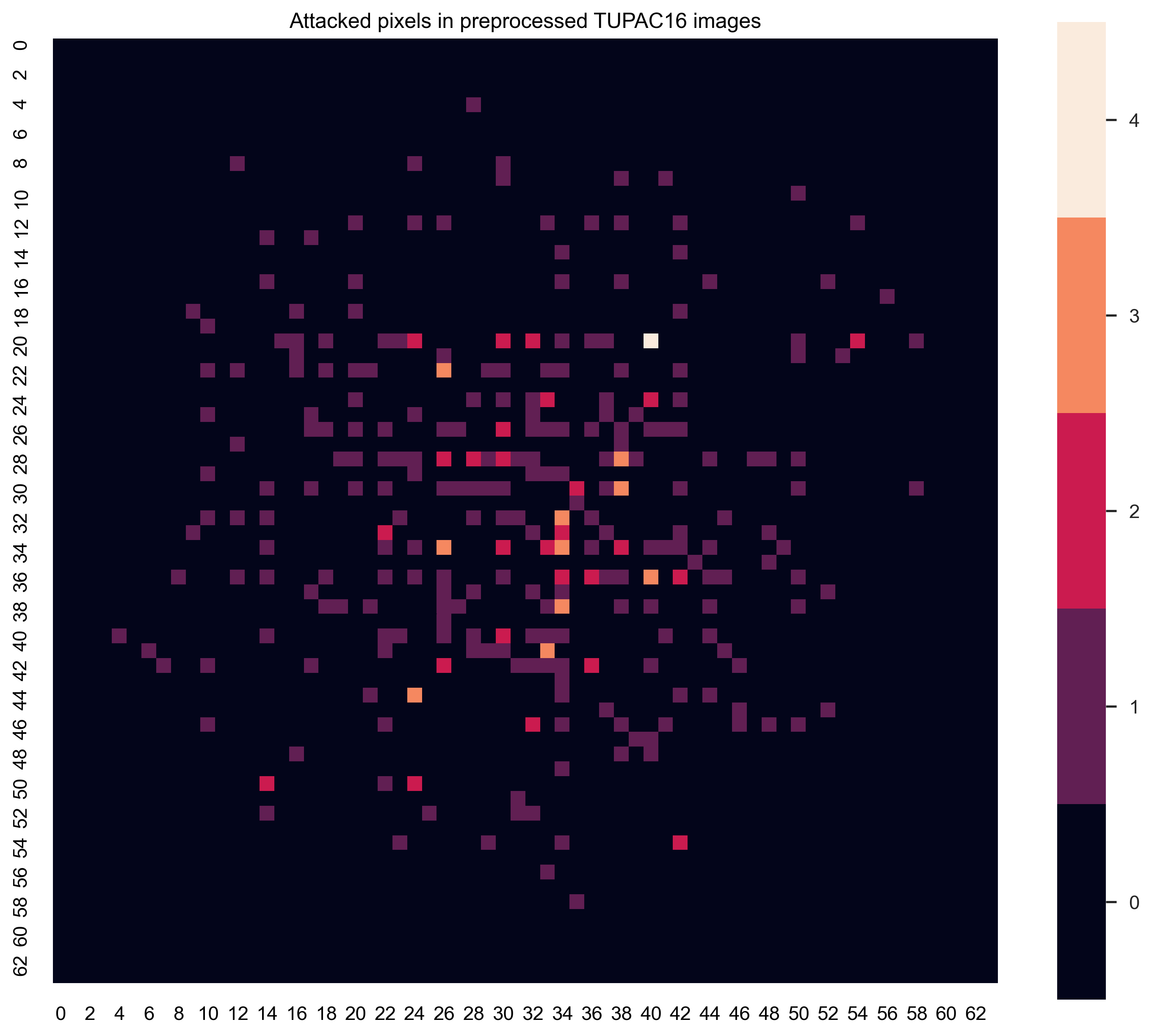}
    \caption{A heatmap of attack locations in images. The attacks are placed mostly around the center of the image.}
    \label{fig:normal-to-mitosis-heatmap}
\end{figure}

There was a smaller ratio of successful attacks in \textit{normal-to-mitosis} direction, and the heatmap visualization in Figure~\ref{fig:normal-to-mitosis-heatmap} does not show any significant clusters or patterns. There is less periodicity and the center of the image is a more prominent location for the successful attacks.

One of the most remarkable features of the spatial diagrams is the periodicity of the \textit{mitosis-to-normal} attacks. Almost all successful attack pixels have coordinates with even numbers. From all of the \(5,343\) \textit{mitosis-to-normal} attacks, the differential evolution algorithm settled on pixel coordinates that had even numbers for both coordinates \(5,334\) times. Only \(9\) times did the algorithm have best success with coordinates where both or one of the coordinates was an odd number. Only \(1\) of the \(9\) odd coordinate attack vectors was successful of lowering the score below \(0.5\) with modified score of \(0.387\).

For \textit{normal-to-mitosis} attacks the coordinates also preferred even coordinates; however, not so clearly. From all of the \(80,725\) attacks \(49,573\) or \(61.4\%\) settled on even coordinates and \(31,152\) or \(38.6\%\) settled on odd coordinates.

Our first reaction was to review the attack code for periodic error but after diligent assessment the code was deemed to be working as it should. This led to the conclusion that a periodic process in the classifier itself was causing this noticeable behavior. The behavior was verified after we brute forced the subset of attack vectors using the method described in~\ref{sec:bruteforce}.

Even with the reduced color space the checkerboard pattern was clearly visible when analyzing the results from the brute force computations. Figure~\ref{fig:successful-m2n-bruteforce-min-heatmap} shows an example image where the minimum confidence score is visualized for each pixel in the image from all the computed attack vectors. As can be seen in the image, the same checkerboard pattern is clearly visible.

The effect of even coordinates being more vulnerable to pixel modifications might be a side effect of the architecture that the targeted model uses. The model source code shows that the model uses convolutional blocks where convolutional layer stride is set to \((2, 2)\). This could cause the checkerboard pattern. If some filter kernel on a convolutional layer that has the stride of \((2, 2)\) is vulnerable to the pixel modification attack, then that effect would be duplicated to every other pixel while the kernel sweeps across the image dimensions while skipping every other coordinate.

\subsection{Brute force confidence map result analysis}

The brute force computations we performed for some handpicked images for both \textit{mitosis-to-normal} and \textit{normal-to-mitosis} images gave more information about the pixel positions for the successful attacks and a possible explanation why all attacks were not successful.

\begin{figure*}[htp]
  \centering
  \begin{subfigure}[t]{1.0\textwidth}
      \centering
      \includegraphics[width=0.54\columnwidth]{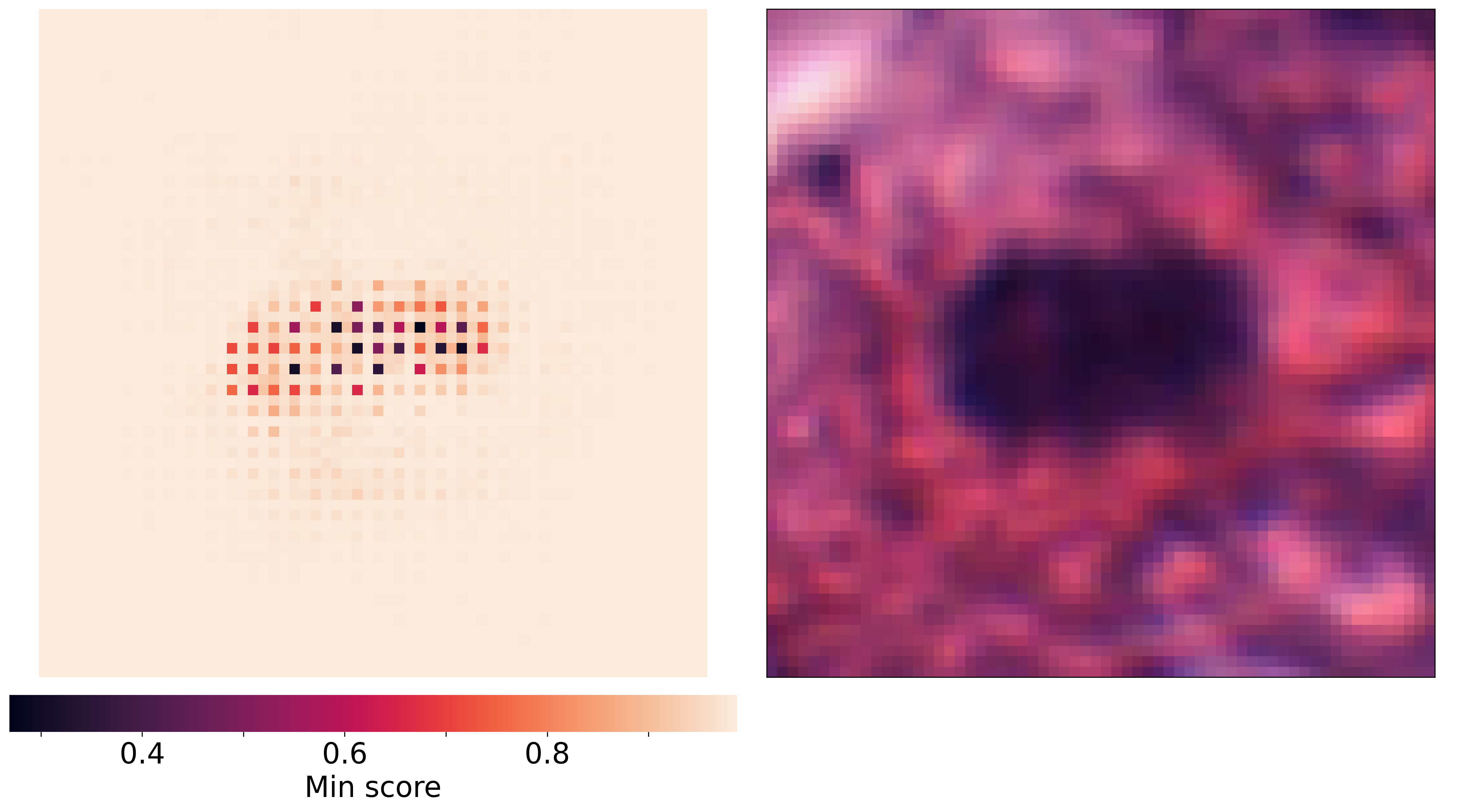}
      \caption{Successful \textit{mitosis-to-normal} attack}
      \label{fig:successful-m2n-bruteforce-min-heatmap}
  \end{subfigure}
  \begin{subfigure}[t]{1.0\textwidth}
      \centering
      \includegraphics[width=0.54\columnwidth]{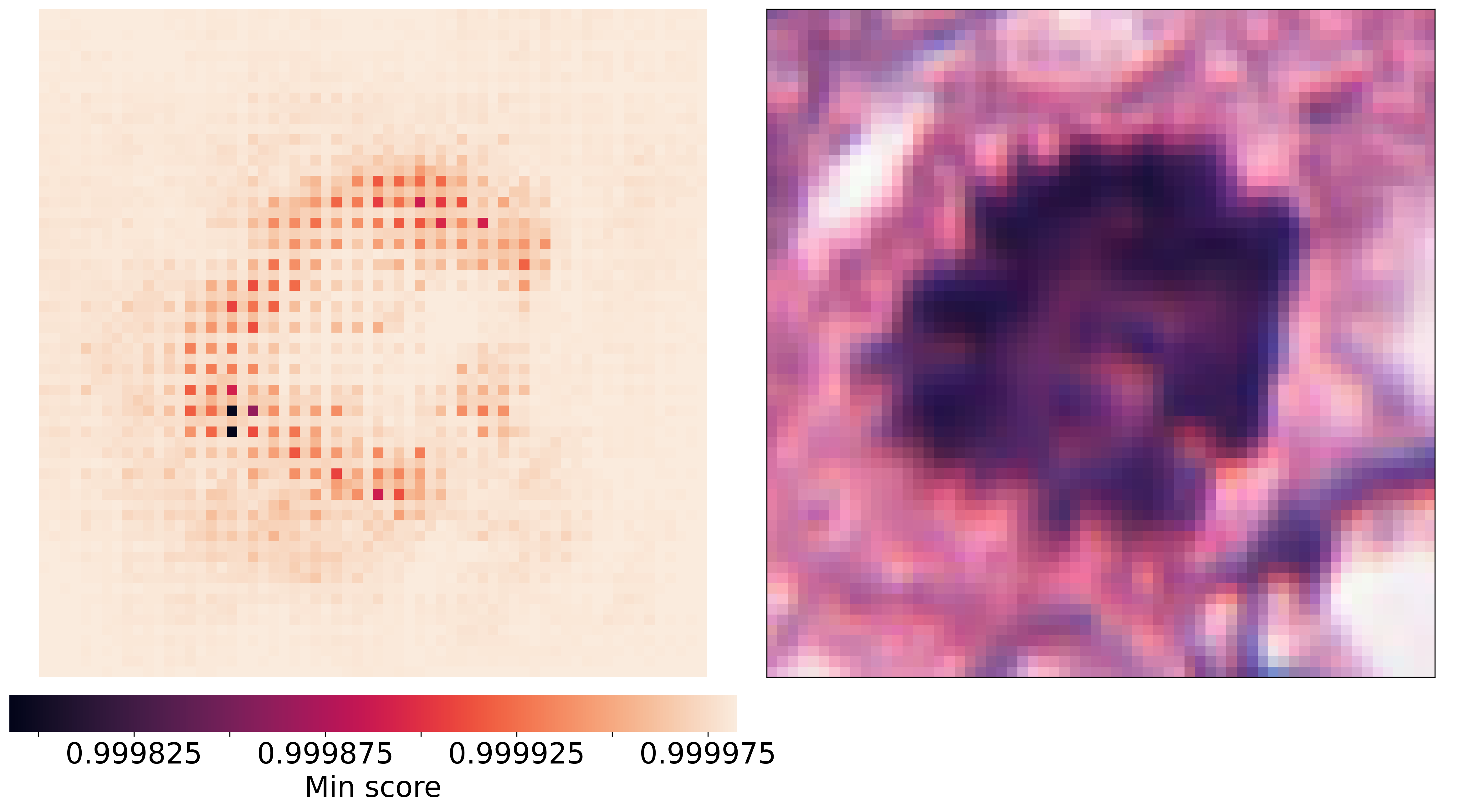}
      \caption{Failed \textit{mitosis-to-normal} attack}
      \label{fig:failed-m2n-bruteforce-min-heatmap}
  \end{subfigure}
  \begin{subfigure}[t]{1.0\textwidth}
      \centering
      \includegraphics[width=0.54\columnwidth]{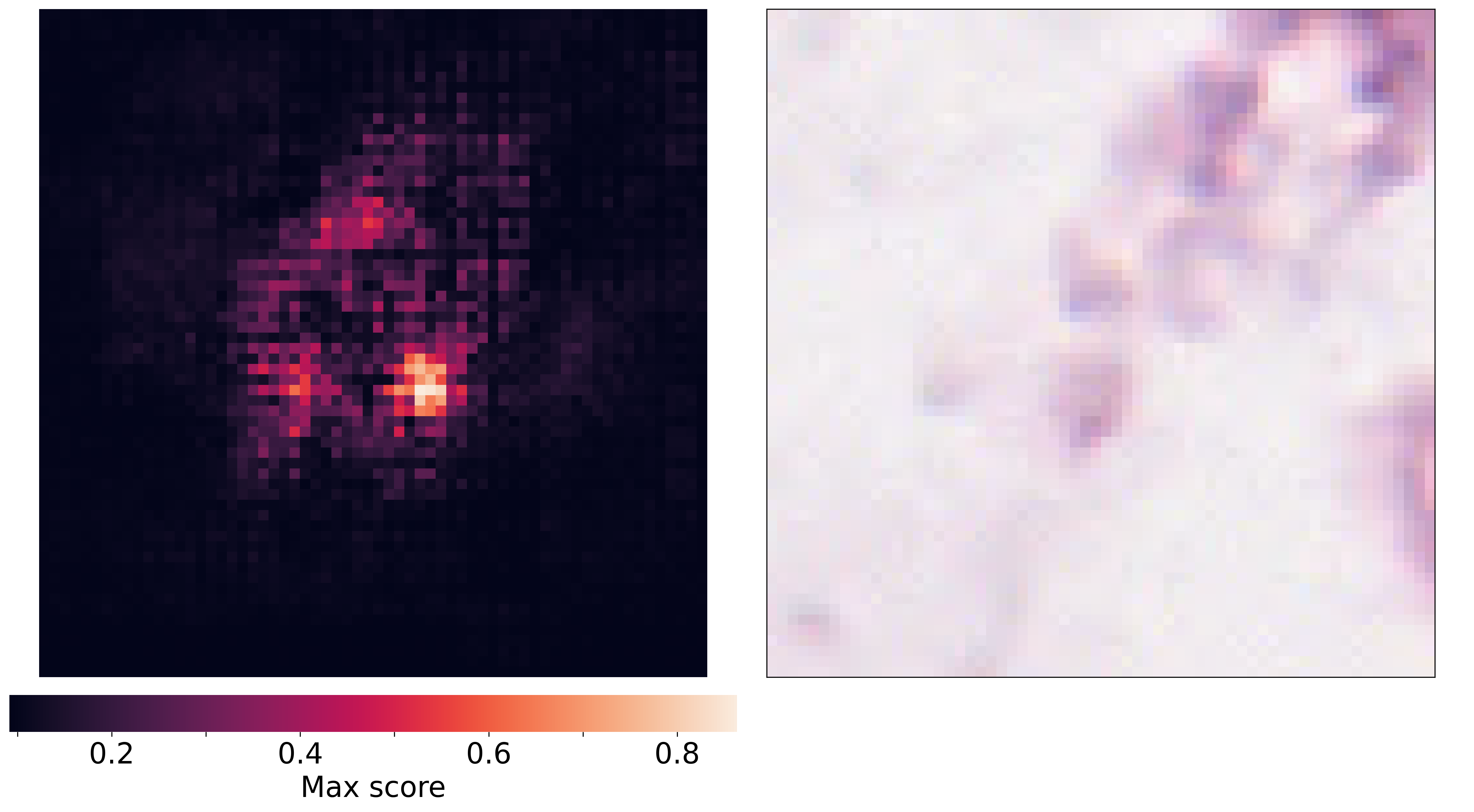}
      \caption{Successful \textit{normal-to-mitosis} attack}
      \label{fig:successful-n2m-bruteforce-max-heatmap}
  \end{subfigure}
  \begin{subfigure}[t]{1.0\textwidth}
      \centering
      \includegraphics[width=0.54\columnwidth]{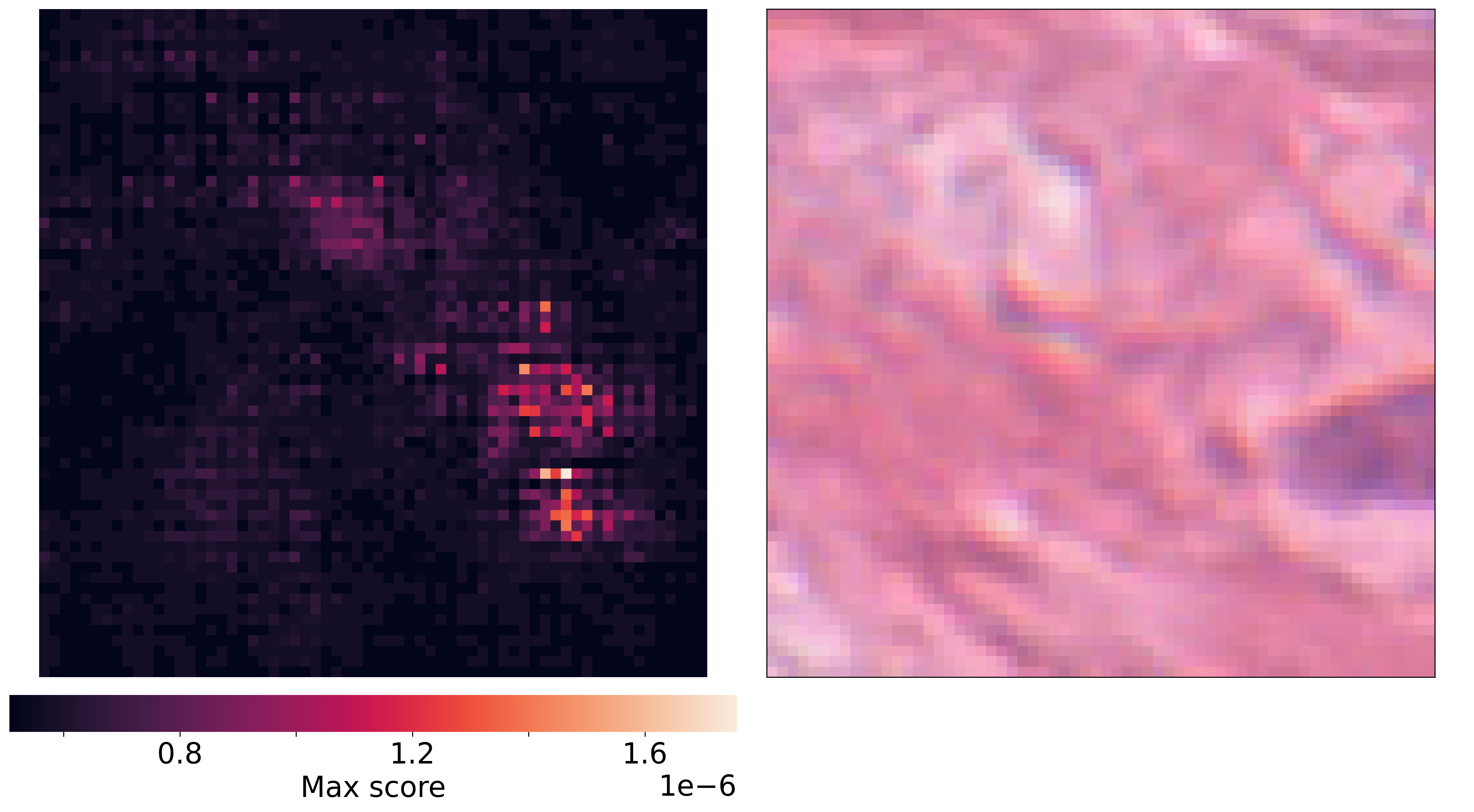}
      \caption{Failed \textit{normal-to-mitosis} attack}
      \label{fig:failed-n2m-bruteforce-max-heatmap}
  \end{subfigure}
  \caption{Images showing examples of successful and failed \textit{mitosis-to-normal} and \textit{normal-to-mitosis} brute force attacks. In each subimage, on the right is the original image under brute force attack, and on the left is a heatmap visualization of maxscores or minscores array that are defined in algorithm \ref{alg:brute-force-processing}. The heatmap image shows the pixel locations that had the biggest impact on the output score when the pixel color was changed.}
  \label{fig:brute-force-subimages}
\end{figure*}

Figure~\ref{fig:successful-m2n-bruteforce-min-heatmap} visualizes the minimum scores for each pixel that were computed for the attack vectors in the executed \textit{mitosis-to-normal} brute force attack for this image. The original score for the image was \(0.9874\) and the lowest score that one of the pixel modifications achieved was \(0.2689\). The image shows that all the attack vectors successfully lowering the score in any meaningful way were situated in the middle of the dark spot in the image.

Figure~\ref{fig:failed-m2n-bruteforce-min-heatmap} shows a similar mitosis image; however, in this image the dark spot is larger. This is an example of a failed \textit{mitosis-to-normal} attack. The original score for this image was \(0.99998\) and the lowest score that any of the attack vectors achieved was \(0.99979\), so the best one-pixel change resulted in practically no change at all. Comparing this image to the successful \textit{mitosis-to-normal} attack in Figure~\ref{fig:successful-m2n-bruteforce-min-heatmap} shows that this time the pixel modifications that were in the middle of the dark spot had absolutely no effect at all, and the pixel modifications that had even the slightest effect to the score were the ones on the edge of the dark spot. This could indicate that the dark spot is so big that the one-pixel modification is not large enough change to fool the model.

Similar to the previously described \textit{mitosis-to-normal} attacks, Figure~\ref{fig:successful-n2m-bruteforce-max-heatmap} and Figure~\ref{fig:failed-n2m-bruteforce-max-heatmap} show successful and failed \textit{normal-to-mitosis} attacks. The successful attack in Figure \ref{fig:successful-n2m-bruteforce-max-heatmap} increased the score from original \(0.09123\) to \(0.86350\), but the failed attack in Figure~\ref{fig:failed-n2m-bruteforce-max-heatmap} had practically no success at all with the original score of \(4.29 \times 10^{-7}\) and the highest achieved score of \(1.04 \times 10^{-6}\). It seems that the successful \textit{normal-to-mitosis} attacks require some kind of dark spots in the middle of the image that the attack pixel highlights by making the spot look bigger and this way fooling the model. If the image does not have a spot in the middle of the image, then one pixel change is not enough to fool the model to think that there is a spot that would indicate a mitosis.

\section{Conclusion}
\label{sec:conclusion}

We have presented a way to systematically analyze the quality of one-pixel attacks. The target images were a set of digital pathology images and the target classifier tried to detect cancerous growth in them. We focused our efforts on the color and location of the attacks, as well as periodicity analysis through confidence maps. 

Chromatic analysis reveals that there are two clusters of attacks. It seems that the confidence score between the original and the adversarial images either stays low or, in the case of successful attacks, gets a rather big boost towards the wanted classification. Furthermore, the attack seems to be more effective the bigger the color difference is. As expected, this creates conflicting multi-objective optimization goals. 

Spatial analysis reveals that the most sensitive areas for the attack are in the middle of the image. This is probably caused by the preprocessing, which produces images that have the prominent feature in the middle. This, in turn, causes the neural network classifier to focus on the middle of the image. Furthermore, combining the spatial and chromatic dimensions, pixels in successful attacks seem to appear inside the dark patches. Another common area is the edge of those dark patches. 

Periodicity analysis shows that some rows and columns are more susceptible to the attack. This stems from the features of the target classification model, which uses a neural network. It seems that a brute force mapping of classifier behavior is useful. The confidence maps illustrate that the most successful attacks are clustered around the dark middle areas of the images. It seems that it is difficult to realize a one-pixel attack if there is no clear dark area. This is caused by what the target classifier is trained to detect, and thus, focus on. 

The methodology presented in this article is suitable for the analysis of any one-pixel attack, and not confined to the world of medical imaging. The only requirement is to have access to a black-box classifier, which produces confidence scores. Such tools should be useful when assessing the quality of the classifier and its robustness. The need of including robustness metrics and mitigation methods to the toolbox of standard implementations seems like the correct direction in future research. 

\subsubsection{Acknowledgments} This work was funded by the Regional Council of Central Finland/Council of Tampere Region and European Regional Development Fund as part of the Health Care Cyber Range (HCCR) project of JAMK University of Applied Sciences Institute of Information Technology.

The authors would like to thank Ms.\ Tuula Kotikoski for proofreading the manuscript.

\bibliographystyle{splncs04}
\bibliography{refs}

\end{document}